\documentclass[letterpaper,10pt,conference]{ieeeconf}

\IEEEoverridecommandlockouts
\usepackage{amsmath,amsfonts,bm}

\def\eqref#1{equation~\ref{#1}}

\def\1{\bm{1}}

\DeclareMathAlphabet{\mathsfit}{\encodingdefault}{\sfdefault}{m}{sl}
\SetMathAlphabet{\mathsfit}{bold}{\encodingdefault}{\sfdefault}{bx}{n}

\usepackage{booktabs}
\usepackage{graphicx}
\usepackage{xcolor}
\usepackage{tikz}
\usetikzlibrary{arrows.meta,calc,fit,positioning}
\usepackage{cite}
\usepackage{url}
\usepackage[hidelinks]{hyperref}

\definecolor{todocolor}{RGB}{150,45,45}

\title{\LARGE \bf
Cycle WAM: Coupling Action and Video through Cycle Consistency
}

\author{%
\authorblockN{Qinglin Yang$^{1,2,\dagger,\ddagger}$, Jiepeng Wang$^{2,\dagger}$, Yuxing Huang$^{3}$, Haibin Huang$^{2}$\\
Kai Xu$^{4}$, Chi Zhang$^{2}$, Ruizhen Hu$^{5,*}$, Xuelong Li$^{2,*}$}
\authorblockA{$^{1}$Zhejiang University \quad
$^{2}$Institute of Artificial Intelligence (TeleAI), China Telecom\\
$^{3}$National University of Defense Technology\\
$^{4}$Institute of AI for Industries, Chinese Academy of Sciences \quad
$^{5}$Shenzhen University\\
$^{\dagger}$Equal contribution. \quad $^{*}$Corresponding authors.\\
$^{\ddagger}$Work done during an internship at TeleAI.}
}
\hypersetup{pdfauthor={Qinglin Yang, Jiepeng Wang, Yuxing Huang, Haibin Huang, Kai Xu, Chi Zhang, Ruizhen Hu, Xuelong Li}}

\begin{document}
\bstctlcite{cyclewam_bib_style}

\IEEEaftertitletext{
\begin{minipage}{\textwidth}
  \centering
  \includegraphics[width=\linewidth]{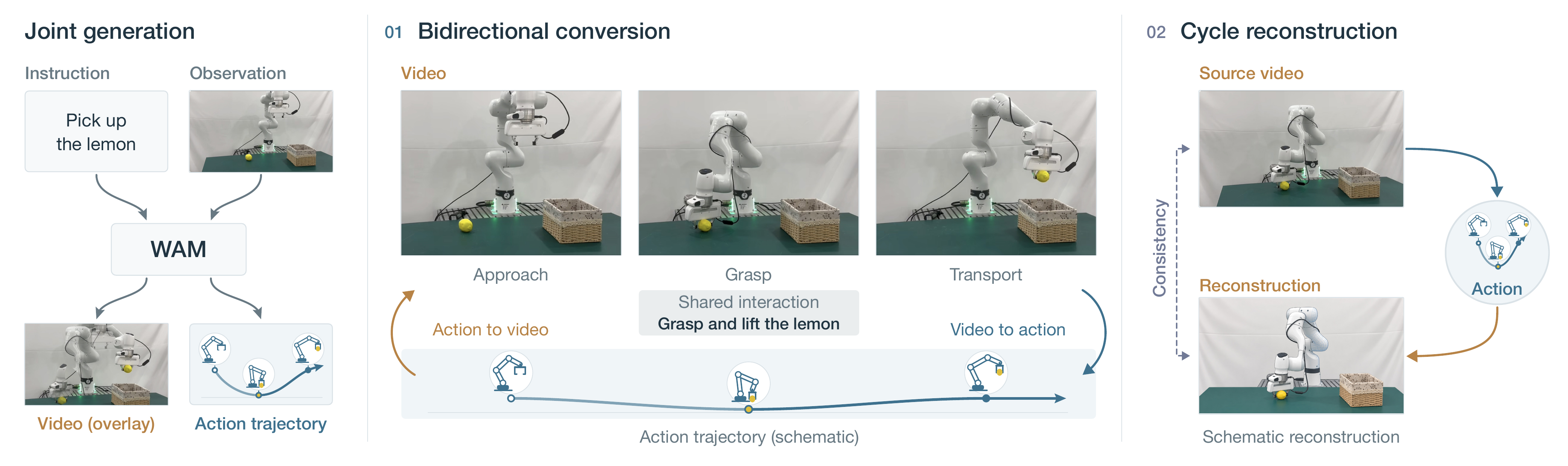}
  \makeatletter
  \def\@captype{figure}
  \makeatother
  \caption{Cycle WAM learns the correspondence between video and action.
  Building on joint generation (left), Bidirectional Conversion relates
  the two modalities through a shared interaction (middle), and Cycle
  reinforces this relationship through reconstruction (right).
  Action trajectories and the reconstruction are schematic.}
  \label{fig:teaser}
\end{minipage}
\par\vspace{0.5\baselineskip}
}
\maketitle
\thispagestyle{empty}
\pagestyle{empty}

\begin{abstract}
%
Recent advances in video modeling have driven substantial progress in world
action models (WAMs).
%
Existing approaches exploit pretrained video priors by decoding actions
from latent representations or inferring them from generated videos.
The underlying link is causal: robot actions produce changes in the scene,
and video observations of these changes provide evidence for inferring the
actions that produced them. However, existing training formulations do not
by themselves ensure that actions and their predicted visual consequences
are consistent, potentially limiting the use of pretrained capabilities.
%
We treat video and action as paired modalities linked by this causal
relationship and learn their correspondence through a unified training
framework designed to better exploit pretrained video priors.
%
Cycle WAM first augments video--action co-training with bidirectional
conversion to reinforce cross-modal correspondence and recover forward and
inverse dynamics. It then introduces a differentiable cycle-reconstruction
objective to further couple the two mappings.
%
The resulting model supports both forward and inverse dynamics, with the
potential to improve performance through more effective use of pretrained
video priors.
%
%
Cycle WAM achieves state-of-the-art performance among comparable methods
on LIBERO-Pro and substantially improves DreamZero on DROID. Comparisons with UWM characterize its inverse-dynamics performance and show
sharper visual detail in the displayed forward-dynamics predictions.
%
\end{abstract}

\section{Introduction}

General-purpose manipulation requires robots to connect desired changes in
the scene with executable actions. Pretrained video models offer rich priors over such changes, motivating world action models (WAMs) that learn visual dynamics and action generation together \cite{uva2025,dreamzero2026,lingbotva2026}. We study how learning the
correspondence between video and action can more fully realize the potential
of pretrained video models for action-conditioned forward dynamics,
inverse dynamics, and policy execution. Within the broader {AI Flow}
framework \cite{an2026ai}, this work focuses on video--action learning
for embodied intelligence.

Early action-conditioned visual models supported model-predictive control
\cite{finn2016interaction,finn2017foresight}, while visual planning
translated predicted states into commands through learned inverse models or
tracking controllers \cite{wang2019visualplanning}. These approaches use predicted visual outcomes to guide action selection, providing a foundation for exploiting richer video priors in robot control.

Recent policies exploit stronger visual priors by conditioning controllers
on generated video plans \cite{unipi2023,gen2act2024} or learning actions
alongside future visual representations \cite{gr12023,gr22024,dreamvla2025}.
These routes transfer visual knowledge, but a plausible future and an action that realizes it remain distinct prediction targets.
%
Building on these advances, unified world action models jointly model robot actions and future visual observations. UWM uses separate diffusion timesteps for video and action to support forward and inverse dynamics \cite{zhu2025uwm}. DreamZero adapts a pretrained video backbone for joint video--action generation \cite{dreamzero2026}.

The connection between video and action is rooted in the causal structure
of an interaction. Given the current scene, robot actions produce changes
that become visible in subsequent observations. Conversely, these visual
changes provide evidence about which actions could have produced them,
enabling inverse dynamics inference even when the underlying action is not
uniquely determined. A shared backbone does not by itself ensure consistency between predicted actions and their visual consequences. Our goal is to strengthen this correspondence so that pretrained video knowledge can support both predicting action outcomes and inferring actions from video.

We therefore view video and action as complementary modalities describing
the same underlying interaction. Bidirectional conversion and cycle
consistency encourage the model's internal representations to capture how
actions relate to visual changes, helping it make fuller use of pretrained
video priors without introducing an explicit intermediate representation.

We introduce \emph{Cycle WAM}, a unified training framework for learning
this correspondence (Fig.~\ref{fig:teaser}). Building on video--action co-training,
\emph{Bidirectional Conversion} trains the model to convert between video
and action, strengthening their correspondence and recovering forward and
inverse dynamics. We further introduce a Cycle mechanism that reconstructs
the input modality after conversion to the other modality, reinforcing the
relationship learned through bidirectional conversion.
%
By learning to convert between video and action and reconstruct the input
modality, the model is encouraged to preserve information about the
interaction that the two modalities share. Our theoretical analysis clarifies how bidirectional conversion and cycle consistency complement each other in learning the correspondence between video and action. Together, these training objectives aim to make pretrained video priors more useful for learning both visual dynamics and robot actions, supporting forward dynamics, inverse dynamics, and policy learning within the same WAM.

Experiments on LIBERO-Pro and DROID demonstrate improvements in model
performance with our training strategy. Comparisons with UWM on LIBERO
further characterize the model's inverse- and forward-dynamics capabilities,
with qualitative forward predictions showing sharper visual detail in the
displayed examples.

Our contributions are:
\begin{itemize}
  \item We study WAM learning from the perspective of video--action
  correspondence to better exploit pretrained video priors, and provide
  theoretical analysis of how bidirectional conversion and cycle
  consistency complement each other in learning this relationship.
  \item We introduce Cycle WAM, a unified training framework that extends
  video--action co-training with Bidirectional Conversion and Cycle
  reconstruction to learn and reinforce the correspondence between the
  two modalities.
  \item We demonstrate performance improvements with our training strategy
  on LIBERO-Pro and DROID, and characterize the model's forward- and
  inverse-dynamics capabilities through comparisons with UWM on LIBERO.
\end{itemize}

\section{Related Work}

\paragraph{Action-conditioned video prediction}
Action-conditioned video prediction models the visual consequences of robot
commands and supports model-predictive control
\cite{finn2016interaction,finn2017foresight}.
Learned simulators extend this approach to robot manipulation
\cite{irasim2024}, while predictive latent models support planning without
pixel reconstruction \cite{vjepa22025}.
Complementary visual-planning methods use predicted futures to guide
a separately learned controller or inverse model
\cite{wang2019visualplanning,unipi2023,gen2act2024}.
These predictors need not also generate robot actions. Beyond robotics,
generative video compression reconstructs video from compact transmitted
representations using generative priors \cite{chen2026generative}.

\paragraph{World action models}
Unified models bring video and action generation into a shared architecture
\cite{zhu2025uwm,uva2025,worldvla2025}.
Related video-based policies couple action learning with future prediction
\cite{gr12023,gr22024,dreamvla2025}.
UWM is a particularly close precedent: independent modality timesteps expose
policy, forward dynamics, inverse dynamics, and video-generation modes
\cite{zhu2025uwm}.
DreamZero, our starting architecture, adapts a pretrained video model for
joint video--action generation \cite{dreamzero2026}.
Recent WAMs extend this design through causal video--action modeling,
visual--action tokenization, and action-centered representations
\cite{lingbotva2026,lingbotva2_2026,repwam2026,actionimages2026,gigaworld2026},
as well as language-conditioned reasoning and broader training data
\cite{wla2026,dswam2026,zerowam2026,motubrain2026,tau0wm2026}.
These architectures differ in how visual priors reach the action prediction.
The broader role of prior knowledge in representation learning is also
emphasized in deep clustering \cite{deepclustering2024}.
Our complementary question is how to train the correspondence between
video and action once both are represented inside a shared WAM.

\paragraph{Coupling forward and inverse dynamics}
The use of forward models to train controllers predates current WAMs.
Distal supervised learning~\cite{jordan1992distal} propagates outcome
errors through a learned forward model, and paired internal models connect
prediction with motor commands~\cite{wolpert1998paired}.
Joint visual forward--inverse learning subsequently used action supervision
to shape predictive interaction features
\cite{agrawal2016poke,pathak2017curiosity}.
More broadly, multimodal generation models connect complementary
descriptions of images, videos, and 3D shapes through unified modeling
or shared representations
\cite{xi2026survey,mmgen2025,omnivdiff2026,ctrlvdiff2025,wang2025structre,dong2025crossgen}.
Cycle consistency provides an explicit round-trip constraint
\cite{cyclegan2017}: D3G uses inverse--forward composition to regularize
reachability \cite{d3g2020}, while forward--inverse cycle consistency
(FICC)~\cite{ficc2023} reconstructs future representations through quantized
latent actions.
WAV uses related cycles for world-model verification and self-improvement;
SC3-Eval jointly trains forward and inverse modes and uses inverse agreement
to detect rollout drift at inference \cite{wav2026,sc3eval2026}.
Cycle WAM builds on this established principle to couple video and action
as two descriptions of an interaction. Bidirectional Conversion supplies paired supervision;
Cycle trains forward dynamics on an action generated by the same WAM's
inverse dynamics, with gradient through that action. Action anchoring and
joint-policy rehearsal connect this compositional objective to robot
coordinates and the deployment query. This is a training-level distinction,
not a claim that unified interfaces or forward--inverse coupling are new.

\begin{figure*}[t]
\centering
\begin{tikzpicture}
 \node[inner sep=0] (art) {\includegraphics[width=\textwidth]{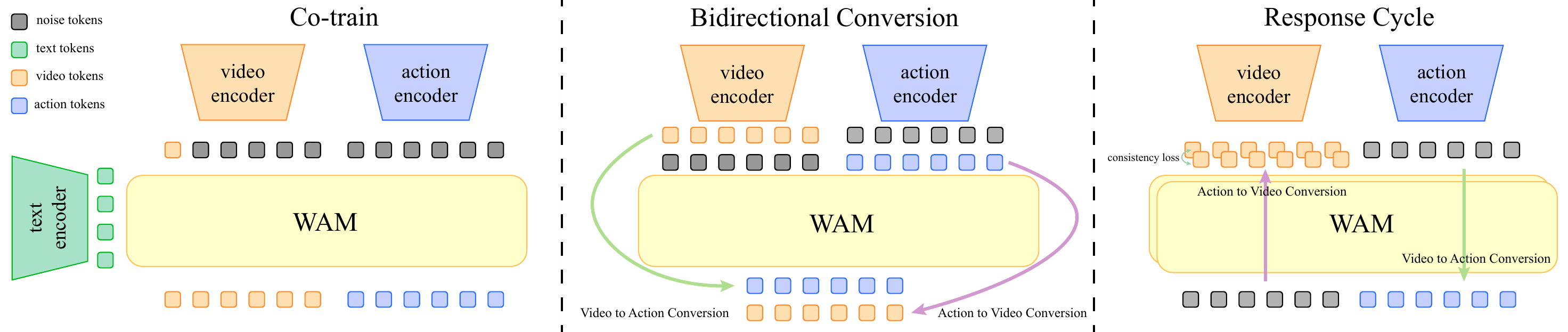}};
 \node[fill=white,minimum width=64pt,minimum height=12pt,inner sep=1pt,
       font=\rmfamily\small] at
       ($(art.north west)!0.861!(art.north east)+(0,-5pt)$) {Cycle};
\end{tikzpicture}
\caption{Overview of Cycle WAM. Left: co-training jointly models future
video and robot actions using a video-pretrained WAM. Middle: Bidirectional
Conversion learns video-to-action (green) and action-to-video (purple)
mappings from paired data. Right: Cycle converts video to action and back
to video, using reconstruction to reinforce the learned correspondence.
The WAM blocks share parameters. Gray tokens denote noisy inputs; orange,
blue, and green tokens denote video, action, and text, respectively.
The diagram depicts the modality-level tasks; reconstruction is trained
with the native video flow-matching objective.}
\label{fig:architecture}
\end{figure*}

\section{Learning Video--Action Correspondence}
\label{sec:theory}

We view video and action as two modalities that express the same underlying
interaction. Their correspondence can be understood through a shared latent
variable that captures the interaction information represented in both
modalities. From this perspective, bidirectional conversion encourages the
model to learn this shared information from paired video--action data,
while Cycle reinforces its preservation through reconstruction. Our analysis
examines how these complementary objectives help the model learn the
correspondence between video and action.

\subsection{The Prediction Cost of Missing Correspondence}
\label{sec:semantic-correspondence}

Let $C$ be context, $Z$ a fixed shared interaction variable, $H$ the model's
input representation, and $Y$ a prediction target. Assume finite second
moments and conditional-mean sufficiency,
$\mathbb E[Y\mid C,Z,H]=m_Z(C,Z):=\mathbb E[Y\mid C,Z]$.
Writing $m_H=\mathbb E[Y\mid C,H]$, define the population prediction risk
$\mathcal R_Y(d,H)=\mathbb E\|Y-d(C,H)\|^2$. It decomposes as
\begin{equation}
 \mathcal R_Y(d,H)=\mathcal R_Z^*+\Delta_Y(H)+\mathcal E_Y(d,H),
 \label{eq:shared-risk-decomposition}
\end{equation}
where $\mathcal R_Z^*=\mathbb E\|Y-m_Z\|^2$,
$\Delta_Y(H)=\mathbb E\|m_Z-m_H\|^2$, and
$\mathcal E_Y(d,H)=\mathbb E\|d(C,H)-m_H\|^2$.

The term $\Delta_Y(H)$ is the extra prediction error caused by missing
shared information relevant to $Y$; improving the head alone cannot remove
it. The final term measures the head's failure to use the available
representation. The sufficiency assumption must hold for the specified
prediction direction, rather than for every visual detail.

\subsection{Risk Control through Conversion and Cycle}

For paired actions $A$ and videos $V$, let $H_V=h(C,V)$,
$B=d_A(C,H_V)$, and $\widehat V=f(C,B)$, with sampling
randomness fixed and no additional source-video input to $f$.
Define pairing error $P_A=\mathbb E\|B-A\|^2$ and reconstruction error
$D_V=\mathbb E\|\widehat V-V\|^2$. Here $B$ is an action output, not $Z$.

Assume fixed theoretical readouts $\psi_A(C,a)$ and $\psi_V(C,v)$ with
Lipschitz constants $K_A,K_V$ over true and predicted inputs.
Their calibration errors are
$\eta_M=\mathbb E\|Z-\psi_M(C,M)\|^2$ for $M\in\{A,V\}$;
these readouts are analytical witnesses, not added network modules.
Let
\begin{equation}
 \begin{aligned}
 U_{\mathrm{BC}}&=(\sqrt{\eta_A}+K_A\sqrt{P_A})^2,\\
 U_{\mathrm{Cycle}}&=(\sqrt{\eta_V}+K_V\sqrt{D_V})^2.
 \end{aligned}
 \label{eq:paired-cycle-certificates}
\end{equation}
For targets satisfying the preceding sufficiency assumption with
$H=H_V$, if $m_Z(C,\cdot)$ is $L_Y$-Lipschitz on its conditional support,
then
\begin{equation}
 \begin{aligned}
 \mathcal R_Y(d,H_V)-\mathcal R_Z^*
 &\le L_Y^2\min\{U_{\mathrm{BC}},U_{\mathrm{Cycle}}\}\\
 &\quad+\mathcal E_Y(d,H_V).
 \end{aligned}
 \label{eq:correspondence-risk-bound}
\end{equation}

\emph{Proof sketch.} Conditional-mean orthogonality gives
Equation~\ref{eq:shared-risk-decomposition}. The conditional-variance
identity and Lipschitz continuity bound $\Delta_Y(H)$ by $L_Y^2$ times the
minimum MSE for estimating $Z$ from $(C,H)$.
Both $\psi_A(C,B)$ and $\psi_V(C,\widehat V)$ are such estimators from
$(C,H_V)$. Conditional-mean optimality and Minkowski's inequality bound
their squared errors by $U_{\mathrm{BC}}$ and $U_{\mathrm{Cycle}}$,
respectively, yielding Equation~\ref{eq:correspondence-risk-bound}.

\subsection{Implications for Cycle WAM}

Paired conversion constrains agreement with observed actions, while Cycle
provides another requirement on information retained across conversion.
The bound connects these objectives to the prediction cost of insufficient
correspondence, supporting our aim of using video priors for robot learning.
It does not guarantee that both bounds are tight or that every training run
improves over its baseline; actual performance also depends on readout error.

For native flow training, the noisy source supplies additional information.
Relating its loss to this guarantee requires high-noise coverage, independent
noise, and stable readouts. Target-video-conditioned risk also does not
establish autonomous policy performance, which we evaluate separately.

\section{Cycle WAM}
\label{sec:method}

Cycle WAM extends video--action co-training with Bidirectional Conversion
and Cycle reconstruction to learn cross-modal correspondence and better
exploit pretrained video priors. As shown in Fig.~\ref{fig:architecture},
a shared WAM progresses from joint generation (left), to paired conversion
(middle), and then to reconstruction through the other modality (right).

\subsection{Video--Action Co-Training}

Using the notation $(C,A,V)$ from Section~\ref{sec:theory}, co-training
jointly predicts future video and robot actions from context $C$, including
visual observations, robot state, and task instruction. Modality-specific
encoders map video, action, and text to tokens, and the video-pretrained
backbone generates video and action from noisy inputs
(Fig.~\ref{fig:architecture}, left). Its native flow-matching objective is
$\mathcal L_J=\ell_J^V+\ell_J^A$, where the two terms supervise video and
action generation, respectively. Positive-incentive noise relates beneficial
perturbations to reduced task complexity \cite{li2022positive}, and its
variational formulation learns noise generators \cite{vpinoise2025}.
Here, we retain the backbone's native flow-matching noise process.

\subsection{Bidirectional Conversion}

Bidirectional Conversion explicitly trains the relationship between paired
modalities (Fig.~\ref{fig:architecture}, middle). Video-to-action conversion
conditions on video to predict actions, learning inverse dynamics;
action-to-video conversion conditions on actions to predict video, learning
forward dynamics. We combine their native flow-matching losses,
$\ell_I^A$ and $\ell_F^V$, with co-training:
\begin{equation}
 \mathcal L_{\mathrm{conversion}}
 =\rho_J\mathcal L_J+\rho_I\ell_I^A+\rho_F\ell_F^V,
 \label{eq:anchor}
\end{equation}
where the $\rho$ terms are query-sampling probabilities. The resulting
model learns both directions of the correspondence while maintaining the
joint-generation task.

\subsection{Cycle Reconstruction}

Cycle reconstructs the source video through video--action--video
conversion (Fig.~\ref{fig:architecture}, right), encouraging preservation
of the shared interaction information discussed in
Section~\ref{sec:theory}.

GenCeption demonstrates that video-generative backbones can be repurposed
for single-pass perception \cite{wang2026genception}. In our cycle, the
video-to-action query obtains a one-step action estimate using the WAM's
native flow readout:
\begin{equation}
 \widehat A=\operatorname{clip}\!\left(
 \epsilon_A-\sigma_{\max}v^I_\theta(\epsilon_A,\tau_{\max};V,C_I),-1,1
 \right),
 \label{eq:inverse-bridge}
\end{equation}
where $\epsilon_A\sim\mathcal N(0,I)$, $v^I_\theta$ is the predicted action
flow, and $\sigma_{\max}$ is the noise scale at the maximum timestep
$\tau_{\max}$. The query observes $V$; $C_I$ retains robot state but omits
language and additional image-feature conditioning. Clipping bounds the
normalized action output.

The action-to-video query then uses $\widehat A$ for reconstruction, trained
with the native video flow loss:
\begin{equation}
 \mathcal L^V_{\mathrm{return}}=
 \mathbb E_\omega\!\left[
 w_\tau\big\|v^F_\theta(X^V_\tau,\tau;C_F,\widehat A)
       -u^V_\tau\big\|^2
 \right].
 \label{eq:return-fm}
\end{equation}
Here $X^V_\tau$ is the corrupted source video, $u^V_\tau$ its target flow,
$w_\tau$ is the scheduler weight, and $\omega$ collects noise and timesteps;
masking and normalization follow the native loss. $C_F$ includes robot
state, the first-frame visual condition, and observed preceding video
blocks, but no language prompt. This trains video flow on a noisy target;
full sampled reconstructions are evaluated separately.

A separate video-to-action computation retains supervision from recorded
pairs. Together with reconstruction, it gives
$\mathcal L_{\mathrm{cyc}}=\mathcal L^V_{\mathrm{return}}+\ell_I^A$, with
unit weights. Thus, Cycle reinforces correspondence while retaining the
paired conversion objective.

The converted action remains in the computation graph: the returned-video
loss differentiates through the forward query and, through $\widehat A$,
the inverse query. Since both queries share the WAM, this couples action
prediction to its visual consequences. The separate supervised inverse
pass keeps the action output tied to recorded robot commands. A Cycle
update therefore contains three computations: inverse conversion, forward
reconstruction, and supervised inverse prediction. The one-step readout
avoids unrolling an entire denoising trajectory, although clipping can
suppress gradients for saturated action components.

\subsection{Training and Inference}

Starting from the co-trained WAM, we train Bidirectional Conversion and
then mix Cycle reconstruction with joint generation:
\begin{equation}
 \mathcal L_{\mathrm{stage2}}
 =(1-\rho_C)\mathcal L_J+\rho_C\mathcal L_{\mathrm{cyc}}.
 \label{eq:stage2}
\end{equation}
Training probabilities and parameter-update settings are given in the
experimental setup. At inference, the two conversion queries provide
forward and inverse dynamics. For robot control, joint generation produces
video and executable actions from context, without requiring a target
future video or an additional cycle at deployment.

\section{Experiments}
\label{sec:experiments}

%
We evaluate control, conditional dynamics, training-stage effects, and
physical execution on LIBERO, DROID, and real-robot tasks.

\subsection{Experimental Setup}
\paragraph{Model and training}
Using RLinf \cite{yu2025rlinf}, we initialize a DreamZero-style 5B WAM
\cite{dreamzero2026} from Wan2.2-TI2V-5B and perform video--action
co-training on LIBERO-40/90 \cite{libero2023} for 27,000 steps.
Bidirectional Conversion then updates all parameters for 5,000 steps with
joint/inverse/forward probabilities $(0.50,0.25,0.25)$. Cycle uses 3,000
steps, joint/Cycle probabilities $(0.70,0.30)$, and shared LoRA adapters
with rank and scaling 32; base weights, action/state interfaces, and query
embeddings remain frozen.

The 5,652 episodes contain 848,036 frame-indexed window starts. At start
$t$, we sample video at $t,t+2,\ldots,t+64$, actions at $t,\ldots,t+63$,
and states at $t,t+16,t+32,t+48$, repeating the final frame at episode
boundaries. Inputs are two camera views, 8D states, and an instruction;
outputs are 7D actions and video latents.

\paragraph{Control evaluation}
We report success rates under the Harness VLA direct protocol on
LIBERO-Pro \cite{liberopro2025}, using 10 tasks and seeds 1--10 per
condition with horizons of 220 steps (520 for LIBERO-10). The selected
four-condition comparison covers 400 episodes; complete coverage includes
800 Pro and 400 Standard episodes. Captions define perturbations and
identify public baseline results.

\subsection{Robot Control on LIBERO-Pro}
Cycle WAM leads the selected-condition average and ranks first on
Spatial-S and Goal-T (Table~\ref{tab:main-results}), supporting control
improvements beyond joint generation. Spatial-T remains below DreamZero
and Goal-S below $\pi_{0.5}$. Cycle WAM also outperforms LIBERO-finetuned
UWM across these conditions. Task, seed, horizon, and success rules are
shared, but UWM's different action chunks and absence of state/language
inputs make this a cross-architecture comparison.
\begin{table*}[t]
\centering
\caption{Success (\%) on four selected LIBERO-Pro conditions.
T denotes instruction redirection and S position swaps. The final column
averages only these four conditions. Baseline values are reported by
LIBERO-Pro~\cite{liberopro2025}, except for the $\dagger$-marked
LIBERO-130-finetuned checkpoints, which we evaluate alongside Cycle WAM
with 100 episodes per condition.}
\label{tab:main-results}
\small
\setlength{\tabcolsep}{8pt}
\begin{tabular}{lrrrrr}
\toprule
Method & Spat-T & Spat-S & Goal-T & Goal-S & Avg. \\
\midrule
OpenVLA~\cite{openvla2024}
  & 0.0 & 0.0 & 0.0 & 0.0 & 0.0 \\
$\pi_0$~\cite{pi02024}
  & 0.0 & 0.0 & 0.0 & 0.0 & 0.0 \\
$\pi_{0.5}$~\cite{pi052025}
  & 1.0 & 20.0 & 2.0 & \textbf{38.0} & 15.3 \\
MolmoAct~\cite{molmoact2025}
  & 0.0 & 0.0 & 0.0 & 0.0 & 0.0 \\
NORA~\cite{nora2025}
  & 0.0 & 0.0 & 0.0 & 0.0 & 0.0 \\
X-VLA~\cite{xvla2025}
  & 0.0 & 0.0 & 9.0 & 1.0 & 2.5 \\
AtomVLA~\cite{atomvla2026}
  & 1.0 & 16.0 & 11.0 & 2.0 & 7.5 \\
\midrule
UWM$^{\dagger}$~\cite{zhu2025uwm}
  & 21.0 & 1.0 & 3.0 & 0.0 & 6.3 \\
DreamZero$^{\dagger}$~\cite{dreamzero2026}
  & \textbf{27.0} & 13.0 & 12.0 & 2.0 & 13.5 \\
\textbf{Cycle WAM (ours)}
  & 24.0 & \textbf{25.0} & \textbf{19.0} & 3.0 & \textbf{17.8} \\
\bottomrule
\end{tabular}
\end{table*}

\subsection{Robot Control on DROID}
Both trained variants outperform official DreamZero-DROID
\cite{dreamzero2026}, with full Cycle WAM best overall (Table~\ref{tab:droid}).
Most improvement follows Bidirectional Conversion; Cycle adds a smaller
gain in the final scene. Unpaired ablations and incomplete training-budget
records limit Cycle attribution. The w/o Cycle variant uses forward/inverse/joint
probabilities $(0.40,0.40,0.20)$; the caption gives other settings.
\begin{table*}[t]
\centering
\caption{Cycle WAM variants on three DreamZero-DROID 14B simulator scenes
\cite{dreamzero2026}, with 20 episodes per scene.
BC denotes Bidirectional Conversion. The w/o Cycle variant uses 5k steps
of rank-32 LoRA BC; omitting both BC and Cycle gives the official weights.
The two ablated runs use the Full20 protocol but are not paired by seed.}
\label{tab:droid}
\small
\setlength{\tabcolsep}{7pt}
\begin{tabular}{lcccc}
\toprule
Variant & Scene 1 & Scene 2 & Scene 3 & Total \\
\midrule
Cycle WAM
  & 20/20 & 19/20 & 20/20
  & \textbf{59/60 (98.3\%)} \\
Cycle WAM w/o Cycle
  & 20/20 & 19/20 & 19/20
  & 58/60 (96.7\%) \\
Cycle WAM w/o BC and Cycle
  & 20/20 & 17/20 & 15/20 & 52/60 (86.7\%) \\
\bottomrule
\end{tabular}
\end{table*}

\subsection{Forward and Inverse Dynamics on LIBERO}
\paragraph{Inverse dynamics}
We compare conditional action prediction on 100 paired windows (34
LIBERO-40, 66 LIBERO-90), sharing $16\times7$ targets, q99 normalization,
seeds, and 16 denoising steps. Official LIBERO-90-pretrained UWM
\cite{zhu2025uwm}, distinct from the policy checkpoint, uses two history
and two endpoint frames with DDIM; Cycle WAM uses nine two-view frames,
state, and FlowUniPC. Both receive observed future video.

Table~\ref{tab:inverse-dynamics} shows lower rotation error and better
motion-direction agreement for Cycle WAM, while UWM has lower aggregate
action MSE. These component-specific strengths do not imply uniform
superiority. The historical DreamZero row uses different windows and a
longer horizon. Inverse prediction receives future video, so its ranking
need not match autonomous control.
\begin{table}[t]
\centering
\caption{Inverse-dynamics summary: total action error, rotation error,
and motion-direction agreement. UWM and Cycle WAM use 100 paired windows.
$\dagger$ marks LIBERO-130-finetuned DreamZero, a historical reference
with different windows and horizon.}
\label{tab:inverse-dynamics}
\small
\setlength{\tabcolsep}{4pt}
\begin{tabular}{@{}lccc@{}}
\toprule
Model & MSE $\downarrow$ & Rot. RMSE $\downarrow$ & Motion cos. $\uparrow$ \\
\midrule
UWM~\cite{zhu2025uwm}
& 0.06 & 0.23 & 0.83 \\
\midrule
DreamZero$^{\dagger}$~\cite{dreamzero2026}
& 0.63 & 0.63 & 0.35 \\
\textbf{Cycle WAM (ours)}
& 0.10 & 0.20 & 0.85 \\
\bottomrule
\end{tabular}
\end{table}

\begin{figure*}[t]
\centering
\includegraphics[width=0.90\textwidth]{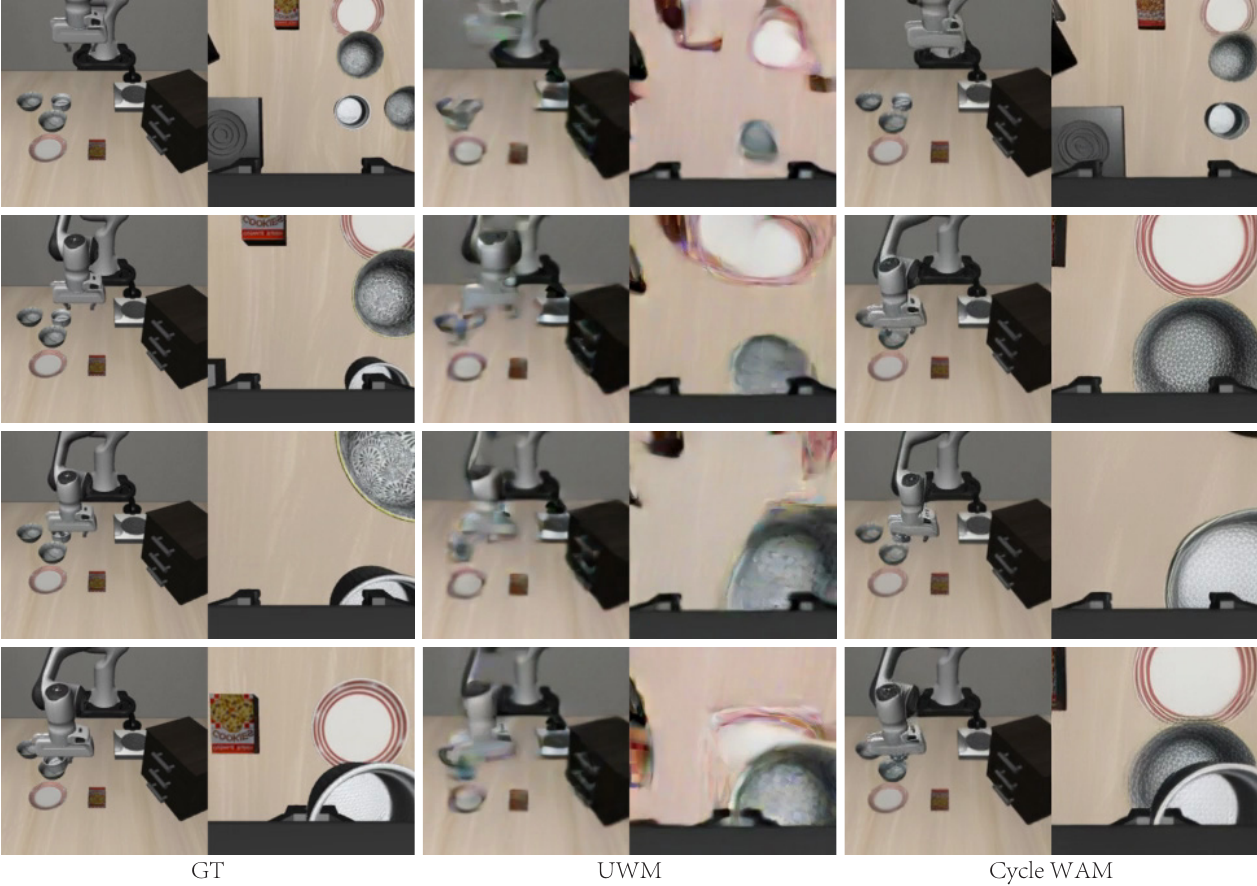}
\caption{Forward-dynamics examples on LIBERO: GT, UWM, and Cycle WAM
(left to right), with two views per panel. Cycle WAM retains sharper
boundaries, but object placement and view composition still differ from GT.}
\label{fig:qualitative}
\end{figure*}

\paragraph{Forward dynamics}
Cycle WAM improves reconstruction error, PSNR, SSIM, and Frame FID
(Table~\ref{tab:forward-dynamics}); its unrounded MSE is 9.65\% lower
despite the displayed tie. Sharper contours and textures in
Fig.~\ref{fig:qualitative} agree with these metrics, although object
placement and view composition differ from GT. This comparison does not
measure temporal consistency or action-intervention sensitivity, or
isolate Cycle's contribution.
\begin{table}[t]
\centering
\caption{Forward-dynamics prediction metrics, rounded to two decimals.
Relative improvements use unrounded scores.}
\label{tab:forward-dynamics}
\small
\setlength{\tabcolsep}{3pt}
\begin{tabular}{@{}lccccc@{}}
\toprule
Model & MSE $\downarrow$ & MAE $\downarrow$ & PSNR $\uparrow$
& SSIM $\uparrow$ & Frame FID $\downarrow$ \\
\midrule
UWM & 0.03 & 0.10 & 15.48 & 0.57 & 280.84 \\
Cycle WAM & 0.03 & 0.08 & 15.92 & 0.63 & 96.02 \\
\bottomrule
\end{tabular}
\end{table}

\subsection{Training-Stage Ablations}
\label{sec:matched-controls}

\paragraph{Training-stage variants}
Bidirectional Conversion improves the selected-condition control average
(Table~\ref{tab:ablation}). Cycle further improves it through Spatial-S and
Goal-T gains, with declines on Spatial-T and Goal-S. Different budgets and
update types prevent isolating each objective's contribution.
\begin{table}[t]
\centering
\caption{Cycle WAM control ablations on four selected LIBERO-Pro conditions.
BC denotes Bidirectional
Conversion; w/o labels indicate omitted training stages. Omitting both
stages gives the LIBERO-finetuned DreamZero baseline \cite{dreamzero2026}.
Success (\%) uses
100 episodes per selected condition. Budgets and update types differ;
matched continuations are discussed in Section~\ref{sec:matched-controls}.}
\label{tab:ablation}
\small
\setlength{\tabcolsep}{3pt}
\begin{tabular}{@{}lccccc@{}}
\toprule
Variant & Spat-T & Spat-S & Goal-T & Goal-S & Avg. \\
\midrule
\textbf{Cycle WAM}
  & 24.0 & \textbf{25.0} & \textbf{19.0} & 3.0 & \textbf{17.8} \\
\quad w/o Cycle
  & 25.0 & 23.0 & 15.0 & \textbf{5.0} & 17.0 \\
\quad w/o BC and Cycle
  & \textbf{27.0} & 13.0 & 12.0 & 2.0 & 13.5 \\
\bottomrule
\end{tabular}
\end{table}


Complete coverage limits these gains: Standard performance falls after
Bidirectional Conversion and is not recovered by Cycle, while aggregate
Pro performance changes little. Control improvements are therefore not
uniform.

\paragraph{Matched continuations}
From the same w/o Cycle checkpoint, we match 3,000-step
rank-32 LoRA runs in data, noise, optimizer, and three-pass computation.
The control replaces converted actions with recorded actions, changing
both conditioning and gradient flow. Runs use 14 or 111 instructions,
10 demonstrations each, seeds 42--44, and a shared 20-task diagnostic
whose separation from training is unverified.

With the larger task set, paired intervals do not resolve differences in
inverse prediction, reconstruction, joint-action consistency, or
action-sensitivity margins. With the smaller set, video reconstruction
and both forward-loss margins favor control under unadjusted intervals;
the others are inconclusive. Matched training thus shows no consistent
additional Cycle benefit and does not separately isolate gradient coupling.
\subsection{Real-Robot Experiments}
\label{sec:real-robot}

\begin{figure*}[t]
  \centering
  \includegraphics[width=0.9\textwidth]{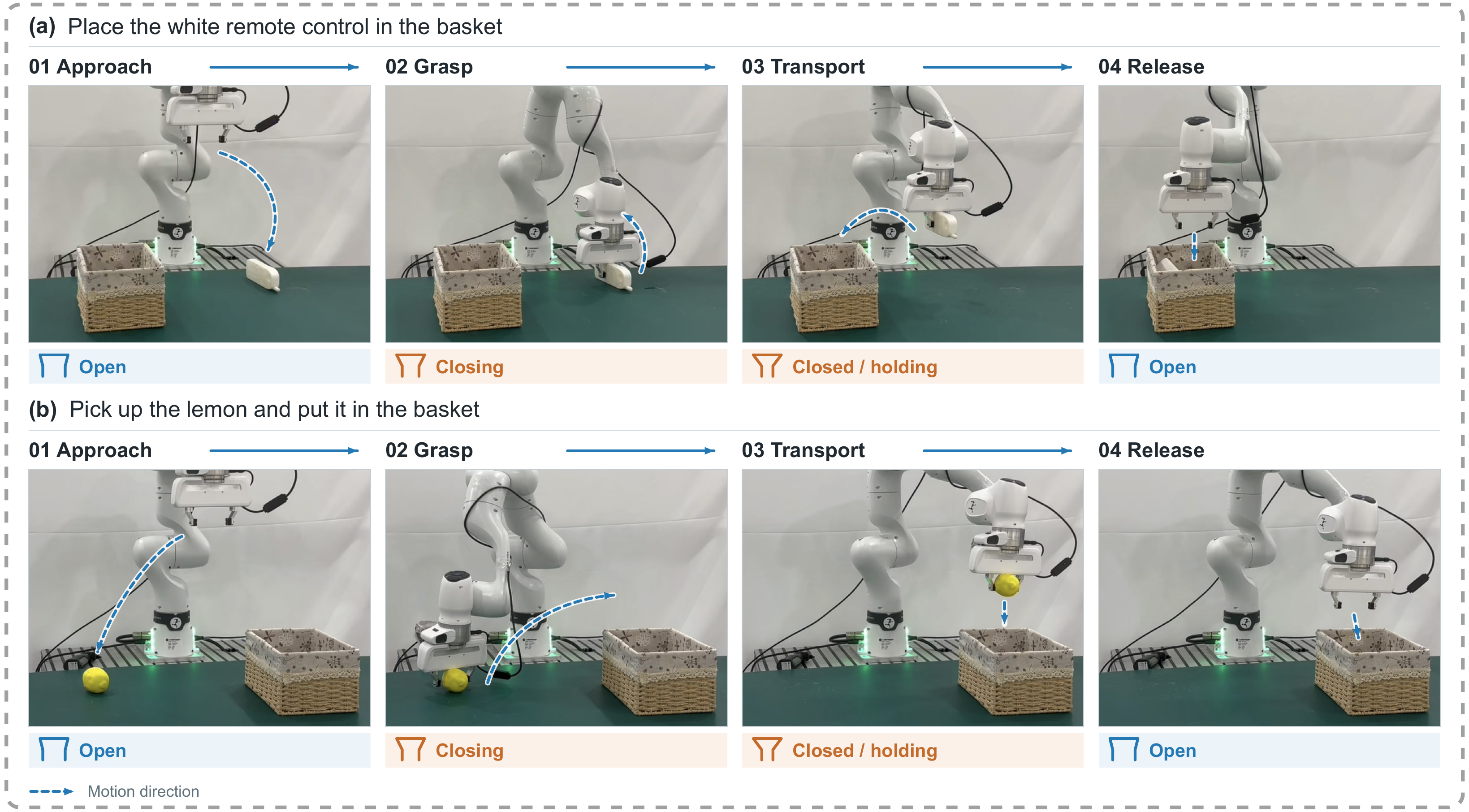}
  \caption{Real-robot placement of (a) a remote control and (b) a lemon
  into a basket. Ordered keyframes show approach, grasping, transport, and
  release. Dashed arrows indicate schematic motion directions; colored
  strips denote visually inferred gripper states.}
  \label{fig:real-robot}
\end{figure*}

Cycle WAM completes the task in three of five evaluated real-world scenes,
whereas DreamZero completes none (Table~\ref{tab:real-robot}). This provides
initial comparative evidence of physical execution, while two scenes
remain unsuccessful for Cycle WAM. These scene-level counts do not establish
repeated-trial reliability.
Fig.~\ref{fig:real-robot} illustrates coordinated arm and gripper motion:
the remote control is placed inside the basket, and the lemon is transported
in the opposite direction and released above it.
\begin{table}[t]
\centering
\caption{Real-robot task completion across five evaluated scenes.
Entries count scenes with observed task completion.}
\label{tab:real-robot}
\small
\setlength{\tabcolsep}{12pt}
\begin{tabular}{@{}lc@{}}
\toprule
Method & Completed scenes \\
\midrule
DreamZero~\cite{dreamzero2026} & 0/5 \\
Cycle WAM & \textbf{3/5} \\
\bottomrule
\end{tabular}
\end{table}


\section{Limitations}


Like other video-pretrained WAMs, Cycle WAM depends on the fidelity and
coverage of its pretrained visual priors. Video-prediction quality can
influence action execution, while precise manipulation and long-horizon
reasoning remain challenging for this model family
\cite{dreamzero2026}. Consistency between video and action cannot by itself
supply missing physical knowledge: the two modalities may agree while
misrepresenting geometry or contact dynamics. The benefit of correspondence
learning is therefore constrained by the underlying model's ability to
represent the interaction. Stronger physical and temporal priors are
important for extending these methods to more demanding manipulation.
Language-guided video segmentation highlights object grounding and
reasoning \cite{lgvos2025}, while training-free long-video understanding
explores frame selection, memory, and reasoning with frozen models
\cite{longvideo2025}. Integrating such capabilities into WAMs remains
an open direction.

A second shared bottleneck is the availability of diverse supervision
beyond RGB video and actions. Acquiring geometric ground truth at scale
remains difficult, motivating synthetic supervision in video-based
perception \cite{wang2026genception}. Extending WAMs to depth, tactile, and
force modalities further requires synchronized and calibrated observations
across tasks and robot platforms. Limited coverage of these signals leaves
aspects of geometry and contact ambiguous from visual observations alone.
Broader multimodal interaction data could provide complementary constraints
on the shared interaction, although collection and alignment costs remain
obstacles to scaling such training.

\section{Conclusion}

Cycle WAM frames video-based robot learning as learning a correspondence
between visual dynamics and robot actions. Bidirectional Conversion learns
from paired modalities, while Cycle reinforces their relationship through
reconstruction. Our analysis explains their complementary roles: paired
supervision identifies the observed correspondence, and reconstruction
encourages information preservation across conversion. Simulation results
characterize bidirectional dynamics and show gains on
selected perturbation conditions, but not uniform improvements in aggregate
control or matched diagnostics. Forward-dynamics comparisons with UWM show
lower reconstruction errors, higher PSNR and SSIM, and lower Frame FID,
alongside sharper visual detail; object-placement discrepancies remain.
Physical-robot tests show task completion in more evaluated scenes than
DreamZero, with qualitative sequences illustrating grasping, transport,
and release in tabletop pick-and-place tasks.
This perspective makes the relation between predicted visual consequences
and executable actions an explicit training and evaluation target.


\section*{Acknowledgment}

We thank Yifei Huang from Nanjing University for assistance with real-robot deployment.


\bibliographystyle{IEEEtranNoURL}
\IEEEtriggeratref{48}
\bibliography{references}



\end{document}